%% file: main.tex
\documentclass[conference]{IEEEtran}
\usepackage{cite}
\usepackage{amsmath,amssymb,amsfonts}
\usepackage{graphicx}
\usepackage[utf8]{inputenc}
\usepackage{textcomp}
\usepackage{xcolor}
\usepackage{subfiles} 
\usepackage{amsmath}
\usepackage{url}
\usepackage{algorithm}
\usepackage{algpseudocode}
\usepackage{subcaption}
\usepackage{setspace} 

\usepackage{booktabs}
\usepackage{amsfonts}
\usepackage{hyperref} 

\def\BibTeX{{\rm B\kern-.05em{\sc i\kern-.025em b}\kern-.08em
    T\kern-.1667em\lower.7ex\hbox{E}\kern-.125emX}}

\title{Generation of Indoor Open Street Maps for Robot Navigation from CAD Files}

\author{
Jiajie Zhang$^{1}$, Shenrui Wu$^{1}$, Xu Ma$^{1}$, S\"oren Schwertfeger$^{1}$
}

\begin{document}
\maketitle
\footnotetext[1]{The authors are with the School of Information Science and Technology, ShanghaiTech University, Shanghai, China. \{zhangjj2023, wushr2023, maxu2023, soerensch\}@shanghaitech.edu.cn}

\begin{abstract}
The deployment of autonomous mobile robots is predicated on the availability of accurate environmental maps. While Simultaneous Localization and Mapping (SLAM) is standard for initial state estimation, maintaining consistent, semantically rich maps in dynamic indoor environments remains a challenge for long-term operations. To address this, this paper presents a complete and automated system for converting architectural Computer-Aided Design (CAD) files into a hierarchical topometric OpenStreetMap (OSM) representation. Instead of replacing on-board sensing, our approach leverages the permanent architectural layout as a static structural prior to enhance navigation robustness. Our core methodology involves a multi-stage pipeline that first isolates key structural layers from raw CAD data and then employs an AreaGraph-based topological segmentation to partition the building layout into navigable spaces. This process yields a comprehensive map, further enriched by a novel heuristic text-to-room association algorithm and an automated multi-floor fusion mechanism that ensures vertical topological connectivity. We validate the system on a diverse dataset of 24 floor plans and, crucially, demonstrate its system-level utility in real-world robotic tasks. Experiments confirm that the generated structural priors effectively support efficient global path planning and robust localization, offering a practical and scalable solution for life-long robot deployments. The software is encapsulated within an intuitive Graphical User Interface (GUI) and is open-sourced at \url{https://anonymous.4open.science/r/osmAG-from-cad-7D01/}.
\end{abstract}


\subfile{sections/introduction}

\subfile{sections/relatedworks}

\subfile{sections/method}

\subfile{sections/experiment}

\subfile{sections/conclusion}

\bibliographystyle{IEEEtran}
\bibliography{references} 

\vspace{12pt}

\end{document}

%% file: sections/introduction.tex
\section{Introduction}
\label{sec:introduction}

The proliferation of autonomous mobile robots in indoor environments, such as hospitals and logistics centers, demands navigation systems that are not only geometrically accurate but also semantically meaningful. For "life-long" autonomy, a robot must possess a map that remains valid despite daily environmental variations, such as moved furniture or dynamic crowds. 
While Simultaneous Localization and Mapping (SLAM) \cite{zhang2014loam} is the standard for estimating robot pose and building metric maps, the resulting representations—typically occupancy grids \cite{moravec1985high} or point clouds \cite{besl1992method}—often lack high-level semantic information and structural permanence. Maintaining these maps over months or years requires costly re-mapping or complex life-long SLAM processes \cite{feng2024fr, mozzarelli2024automatic} to distinguish between temporary obstacles and permanent structures. Furthermore, a significant "format gap" prevents robots from directly utilizing the definitive building data available in Architecture, Engineering, and Construction (AEC) industries.

\begin{figure}[h!]
    \centering
    \begin{subfigure}[b]{0.48\linewidth}
        \includegraphics[width=\linewidth]{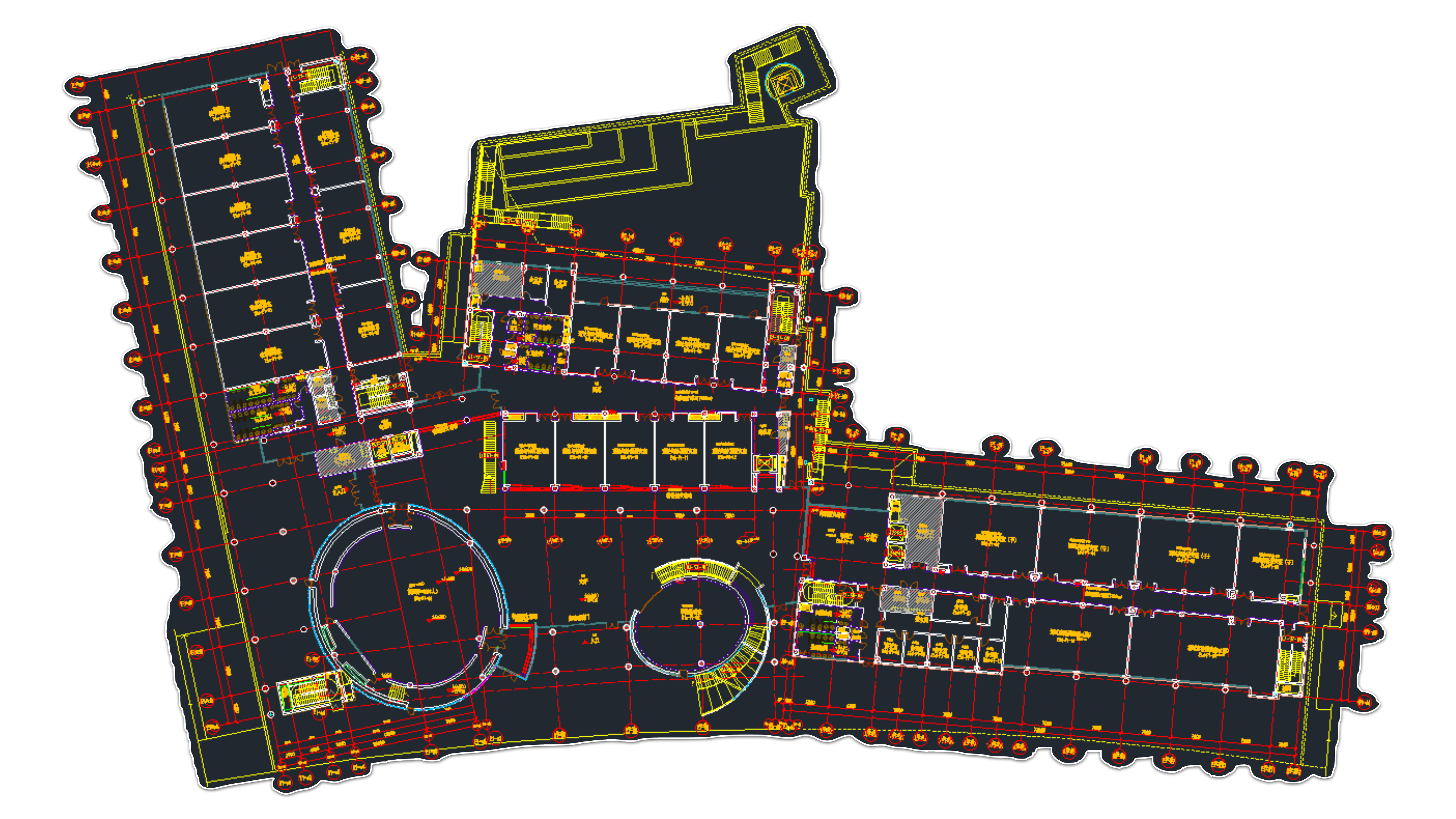}
        \caption{Original Architectural CAD}
        \label{fig:teaser1}
    \end{subfigure}
    \hfill
    \begin{subfigure}[b]{0.48\linewidth}
        \includegraphics[width=\linewidth]{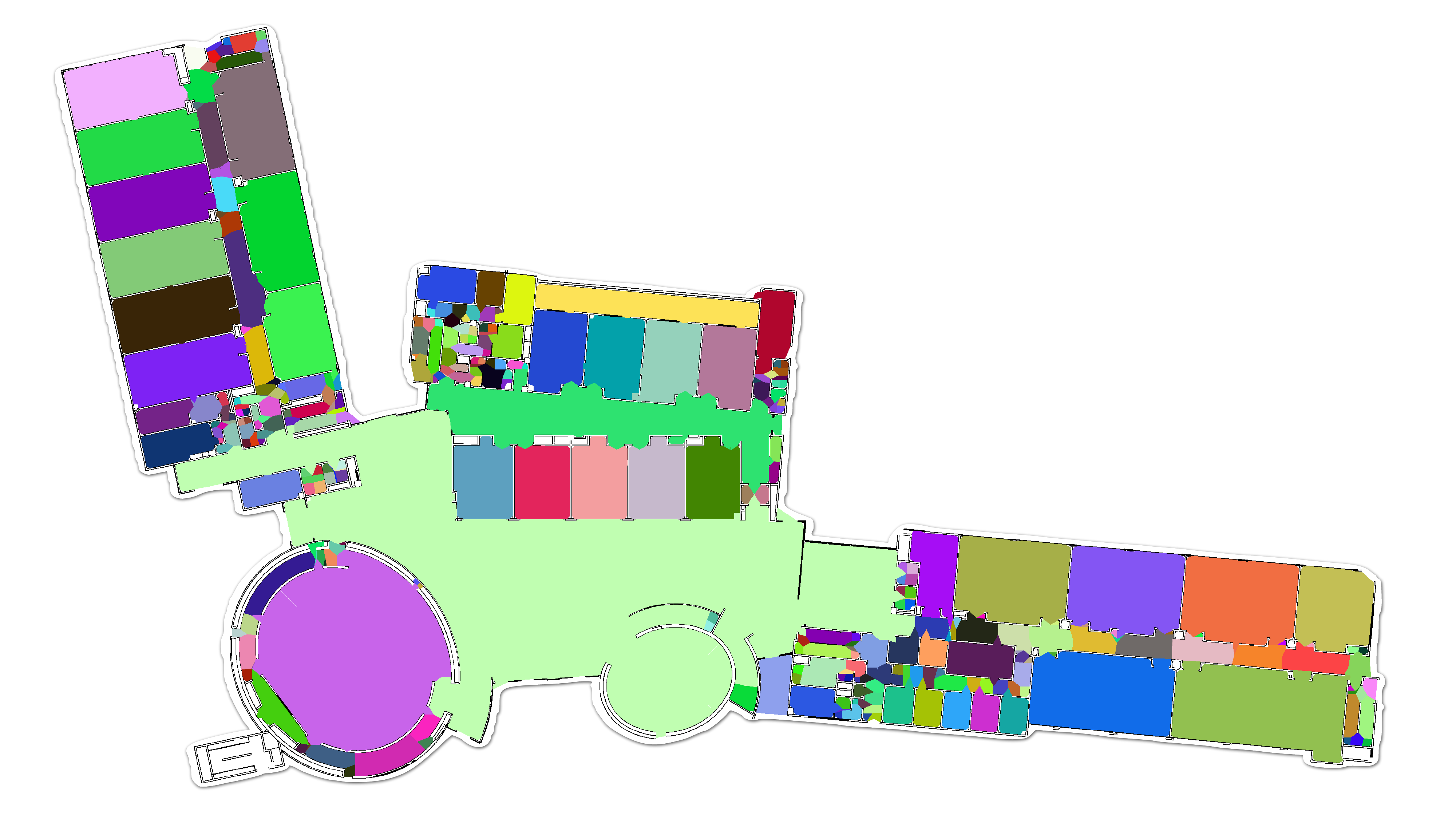}
        \caption{Topological Segmentation}
        \label{fig:teaser2}
    \end{subfigure}
    
    \begin{subfigure}[b]{0.48\linewidth}
        \includegraphics[width=\linewidth]{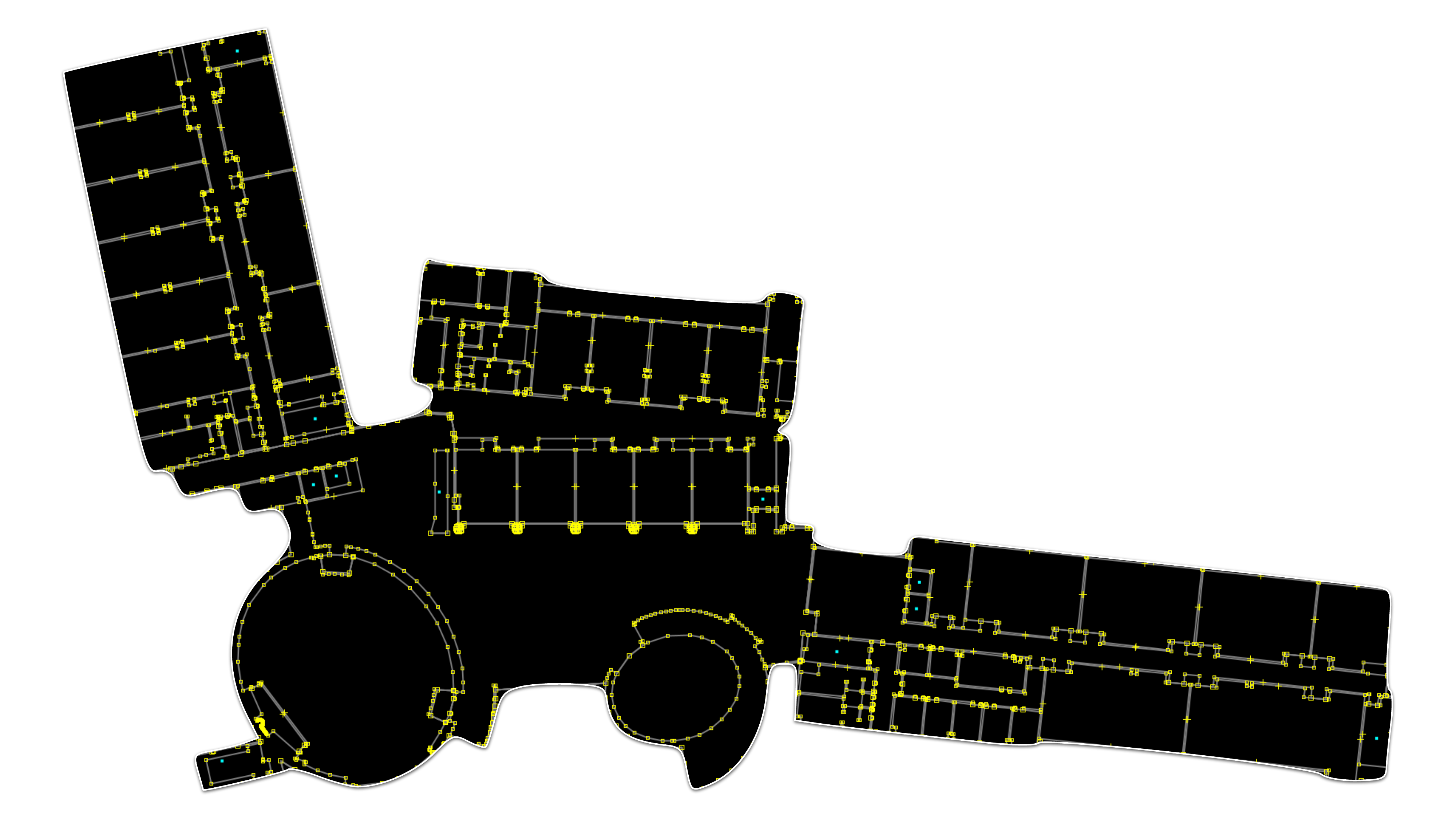}
        \caption{Enhanced OSM in JOSM}
        \label{fig:teaser3}
    \end{subfigure}
    \hfill
    \begin{subfigure}[b]{0.48\linewidth}
        \includegraphics[width=\linewidth]{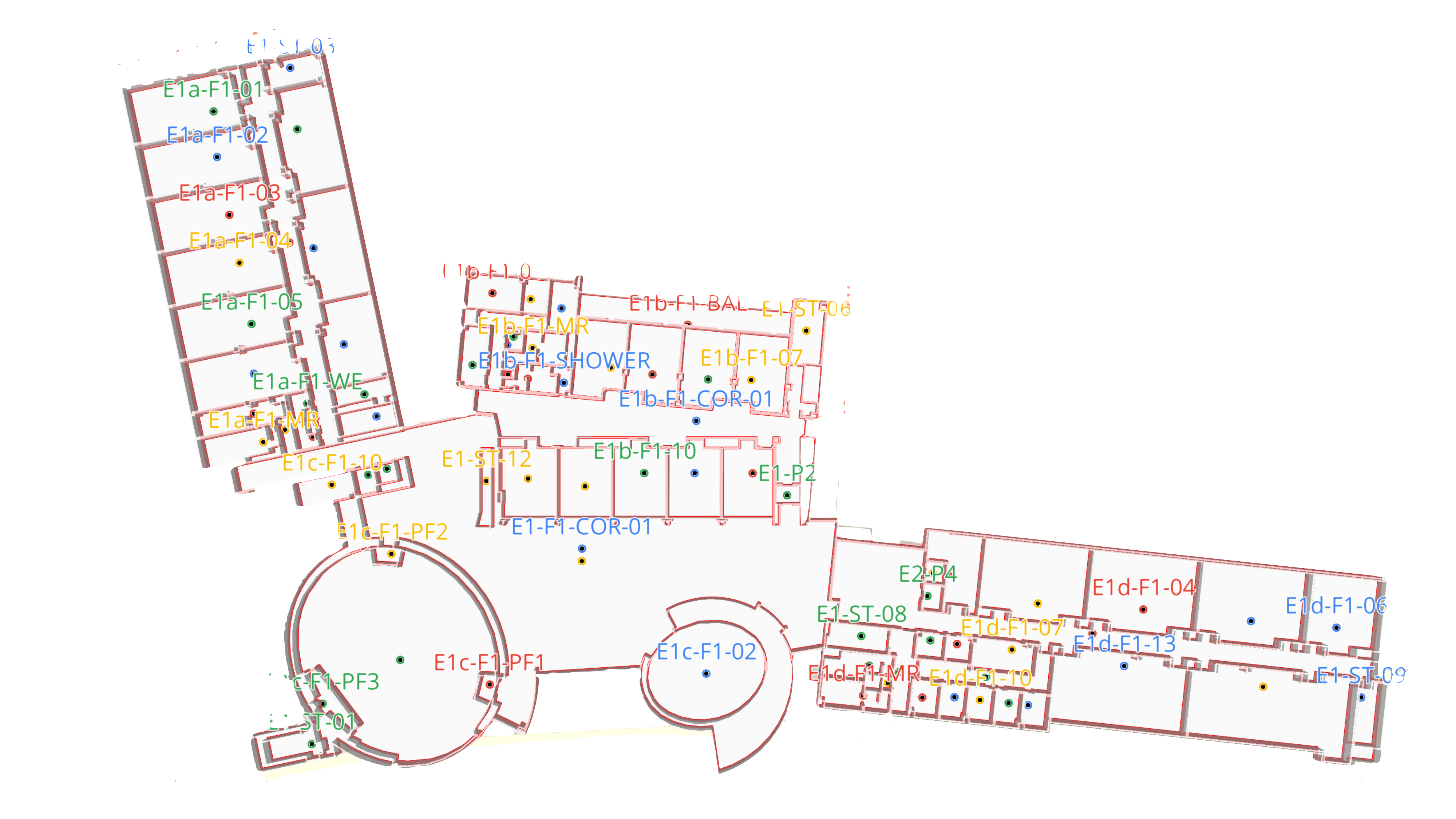}
        \caption{Rendered in OpenIndoor}
        \label{fig:teaser4}
    \end{subfigure}
    \caption{The automatic generation pipeline showcasing key outputs. (\subref{fig:teaser1}) The original architectural CAD drawing, containing numerous non-essential layers and elements. (\subref{fig:teaser2}) The result after topological segmentation into an Area Graph, which initially exhibits over-segmented polygons. (\subref{fig:teaser3}) The refined Area Graph exported to the OpenStreetMap format and visualized in JOSM\cite{osm_wiki:JOSM}; this stage includes a post-processing step where small, insignificant areas are pruned or merged. (\subref{fig:teaser4}) A visualization of our enhanced OSM rendered by OpenIndoor\cite{osm_wiki:openindoor}, demonstrating that our generated map is fully compatible with standard OSM tools.}
    \label{fig:teaser}
\end{figure}

To bridge this gap, we present an end-to-end automated pipeline that transforms raw architectural CAD files into a Hierarchical Topometric Map adhering to the OpenStreetMap (OSM) standard \cite{haklay2008openstreetmap}. We advocate for using this CAD-derived map as an authoritative structural prior. Instead of replacing on-board sensing, our map provides a robust, pre-computed coordinate frame and topological graph that enhances navigation reliability. 

Distinct from recent Deep Learning-based floor plan analysis methods~\cite{liu2017raster, zeng2019deep} which may suffer from generalization issues or hallucinate geometry, our approach utilizes a deterministic, rule-based algorithm. This ensures that the generated map faithfully respects the architect's geometric intent without requiring extensive training data.

The primary contributions of this paper are as follows:
\begin{itemize}
    \item We design an end-to-end system that transforms raw architectural CAD drawings into hierarchical topometric maps compliant with the OpenStreetMap standard~\cite{feng2023osmag}. This establishes a generic data bridge between the Architecture (AEC) and Robotics domains, solving the interoperability bottleneck.
    
    \item We propose a robust topological construction method that integrates dynamic $\alpha$-shape segmentation for scale-invariant horizontal partitioning and an automated multi-floor fusion mechanism for vertical connectivity. This pipeline effectively utilizes geometric and (optional) semantic priors to ensure topological correctness.

    \item We demonstrate the scalability of our approach on a large-scale ($\approx$ 11,025 $m^2$) campus and validate its utility in real-world robotic tasks. Experiments confirm that the generated maps serve as authoritative structural priors, enabling computationally tractable global planning and drift-free localization in dynamic environments.
    
\end{itemize}

%% file: sections/relatedworks.tex
\section{Related Works}
\label{sec:related_works}

Integrating architectural priors provides a static structural backbone for persistent autonomy, overcoming the degradation issues of dynamic SLAM maps~\cite{boniardi2019robot}. Research in this domain has evolved from geometric alignment to semantic parsing and standardization.

\subsection{From Geometric Primitives to Semantic Graphs}

Early methods focused on metric constraints, aligning LiDAR scans to CAD segments~\cite{boniardi2019pose} or synthetic point clouds from BIM~\cite{kaneko2023point,yin2022towards} for localization. While geometrically effective, these approaches lack the semantic context required for high-level tasks. Consequently, research shifted towards topological representations, such as "Architectural Graphs"~\cite{shaheer2023robot,shaheer2023graph} and BIM-derived path planning graphs~\cite{karimi2021semantic}, to fuse structural data with sensor streams.

More recently, Deep Learning (DL) methods like \textit{Raster-to-Vector}~\cite{liu2017raster} and \textit{Deep Floor Plan}~\cite{zeng2019deep} have shown promise in parsing floor plan images. However, these data-driven approaches require extensive training datasets and pose risks of "geometric hallucinations" (inferring non-existent walls), which are unacceptable for safety-critical navigation where deterministic guarantees are preferred. Furthermore, both graph-based and DL approaches often rely on proprietary output formats, hindering interoperability. In contrast, our rule-based pipeline avoids probabilistic uncertainty and outputs directly to the universal OSM standard, ensuring immediate tool compatibility.

\subsection{Standardized Indoor Maps via OpenStreetMap}

While OpenStreetMap (OSM) is the \textit{de facto} standard for outdoor robotics~\cite{hentschel2010autonomous,munoz2022openstreetmap,przewodowski2024global,bashkanov2019exploiting}, its indoor adoption is stalled by the manual labor of digitization. Unlike workflows relying on data-rich but rare BIM/IFC files~\cite{shaheer2023graph,karimi2021semantic}, our work bridges the gap between ubiquitous CAD files and the OSM standard. We automate the generation of \textit{osmAG}~\cite{feng2023osmag}, a hierarchical representation that integrates metric accuracy, topological connectivity, and semantics. By adhering strictly to the OSM XML schema, we transform drawings into living digital assets that support precise localization~\cite{xie2023robust} and can be maintained via standard community tools (e.g., JOSM~\cite{osm_wiki:JOSM}), solving the "closed ecosystem" problem of prior art.

%% file: sections/method.tex
\section{System Methodology}
\label{sec:methodology}
\providecommand{\blind}[1]{{}}

\begin{figure}[t]
    \centering
    \includegraphics[width=0.9\columnwidth]{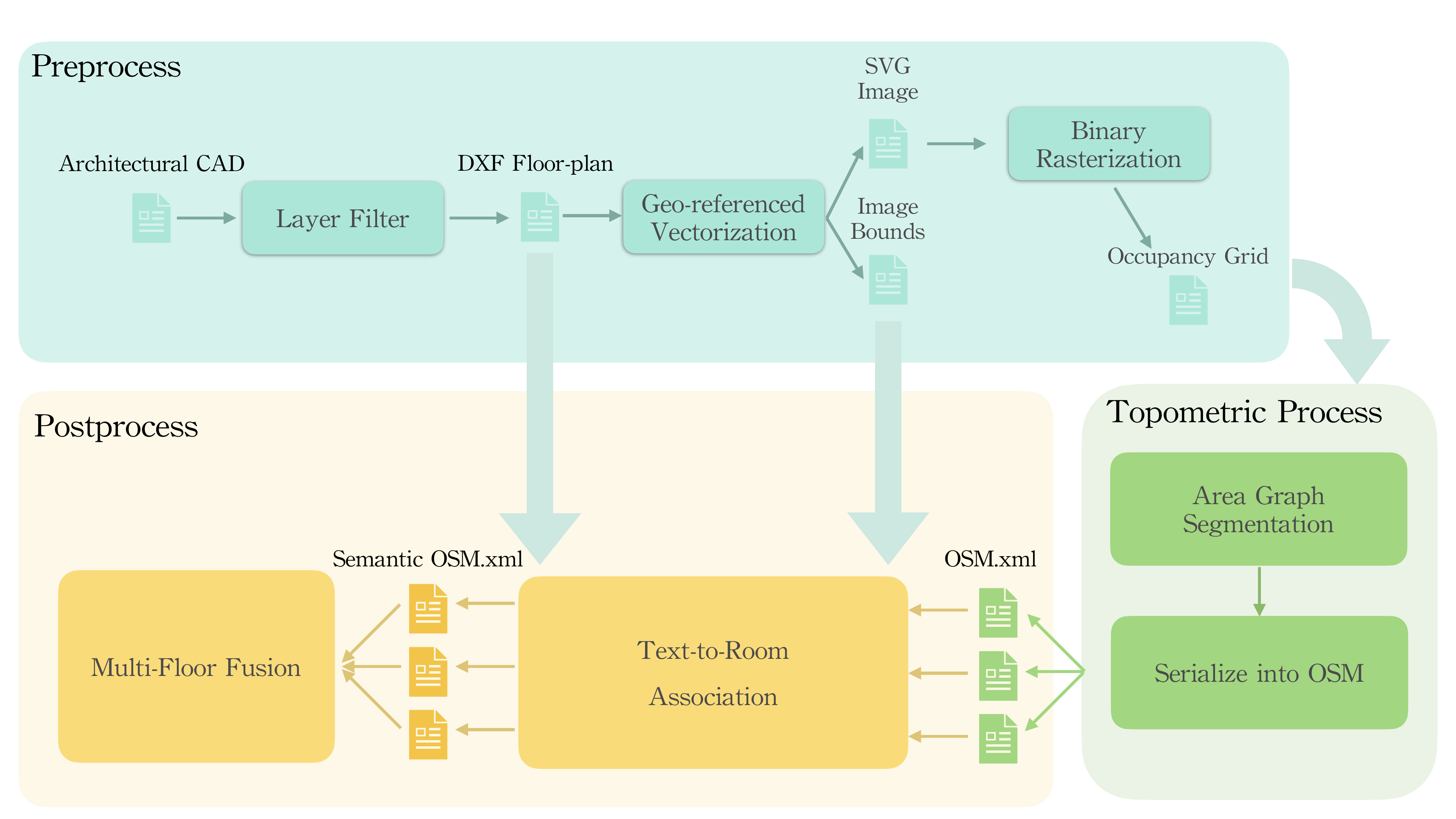}
    \caption{An overview of the proposed map generation pipeline. The system takes a raw CAD floor plan as input and sequentially processes it through stages of preprocessing, topological segmentation, refinement, serialization, and semantic attribution, culminating in a multi-story, enhanced OSM map ready for robotic navigation.}
    \label{fig:pipeline}
\end{figure}

The proposed system systematically transforms architectural CAD floor plans into hierarchical topometric OpenStreetMap (OSM) maps tailored for robotic navigation. An overview of our multi-stage pipeline is depicted in Fig. \ref{fig:pipeline}. The process begins with CAD data preprocessing to generate a clean geometric representation, followed by topological segmentation using an AreaGraph\cite{hou2019area} structure. This graph then undergoes geometric and logical refinement before being serialized into an enhanced OSM format. Finally, semantic information from the CAD source is attributed to the map, and individual floors are fused into a cohesive multi-story model.

\subsection{CAD Preprocessing and Semantic-Aware Rasterization}
\label{subsec:cad_preprocessing}

The initial stage converts the raw architectural CAD floor plan into a standardized binary occupancy grid. This conversion involves three key steps:

\subsubsection{Rule-based Geometry Extraction}
First, we isolate permanent structural features (e.g., walls, columns) using a keyword-based filtering mechanism. Our system targets standard architectural layers (e.g., NCS~\cite{spangler2023c} "A-WALL", “A-STAIR") while strictly preserving reference layers such as "0" and "Defpoints". To ensure robustness against naming variations, the filter is case-insensitive and configurable via external JSON definitions, allowing adaptation to non-standard drawings.

\subsubsection{Adaptive Bounds Computation}
Unlike methods that rely on the file header, we compute the drawing bounds by analyzing the geometry directly. We utilize the $0.5^{th}$ and $99.5^{th}$ percentile of all vertex coordinates to filter outliers (floating debris), adding a $3\%$ safety padding to define the precise physical dimensions. This ensures that the rasterized grid focuses solely on the relevant building footprint.

\subsubsection{Morphological Structure Repair}
During rasterization at a resolution of $r=0.044$ m/pixel, raw vector data often yields disconnected wall segments. We implement a specific \textit{Closing-then-Opening} morphological pipeline using a rectangular kernel $K_{5\times5}$ to watertight the geometry:
\begin{enumerate}
    \item \textit{Gap Bridging:} A morphological \textbf{Closing} operation ($G \bullet K$) connects disjoint wall segments and bridges small gaps ($\approx 0.2$m) caused by drawing errors.
    \item \textit{Noise Removal:} A subsequent \textbf{Opening} operation ($G \circ K$) removes isolated pixel artifacts and smooths jagged edges.
\end{enumerate}
This multi-stage processing produces a clean, closed occupancy grid, which is a vital prerequisite for robust topological segmentation.

\subsection{Topological Graph Generation and OSM Transformation}
\label{sec:areagraph_transformation}

The proposed methodology builds upon the \textit{AreaGraph} formalism~\cite{hou2019area} to partition the free space into a topological graph. Crucially, we enhance the original algorithm with dynamic parameter adaptation to handle varying architectural scales.

\subsubsection{Voronoi Skeletonization}
We generate the Generalized Voronoi Diagram (GVD) of the free space. To eliminate artifacts from furniture or wall textures, the GVD is pruned using a physical distance threshold. Skeleton branches shorter than $d_{prune} = 0.25$m are identified as noise and removed.

\subsubsection{Dynamic $\alpha$-Shape Segmentation}
To prevent over-segmentation in large halls while preserving distinct corridors, we employ an $\alpha$-shape algorithm where the parameter $\alpha$ is not fixed but computed \textit{dynamically}. Specifically, we define $\alpha$ as the \textbf{squared pixel-radius} of a theoretical probe circle derived from physical architectural priors:

\begin{equation}
\label{equation4alpha}
\alpha = \left\lceil \frac{(\min(w_{door}, w_{corridor}) + \delta)^2}{4 \cdot r^2} \right\rceil
\end{equation}
where $w_{door}$ (default 2.8m) and $w_{corridor}$ (default 8.0m) represent typical structural widths, and $\delta=0.1$m is a small adjustment buffer. 
The term implies $\alpha \approx (W/2r)^2$, ensuring that the segmentation kernel adapts to the physical scale of passages regardless of image resolution.

\begin{figure}[t]
    \centering
    \includegraphics[width=\linewidth]{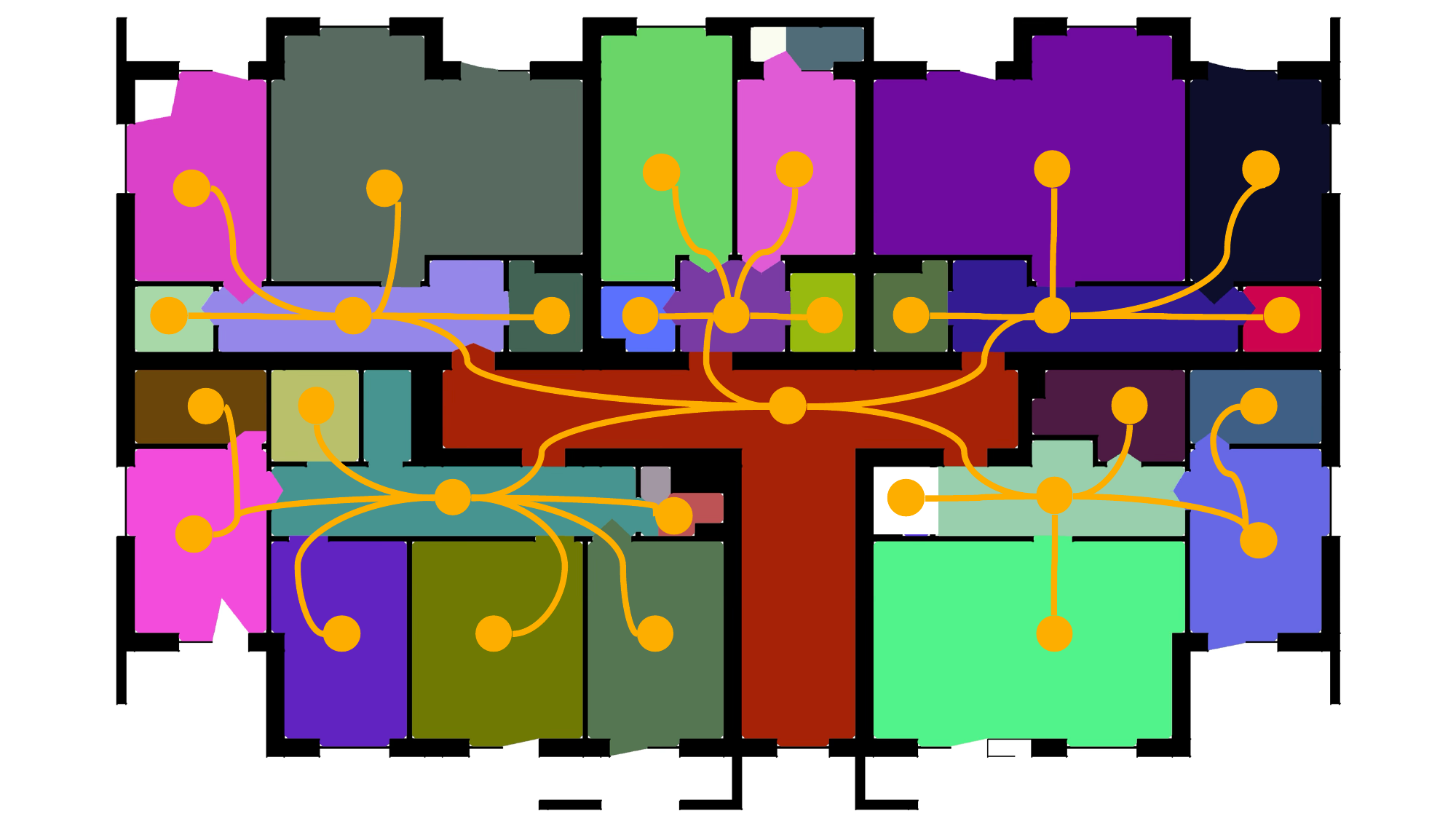}
    \caption{Illustration of the AreaGraph topological structure derived from a floor plan. Nodes correspond to segmented polygonal areas (e.g., rooms, corridors), and edges represent the passages connecting them, forming the foundational topological map for navigation.}
    \label{fig:areagraph_structure}
\end{figure}

\subsubsection{Geometric Simplification with Physical Bounds}
The raw AreaGraph polygons derived from rasterization contain dense pixel-level vertices. To optimize storage without compromising geometric fidelity, we apply a two-stage refinement process:
\begin{enumerate}
    \item \textbf{Adaptive Simplification:} We utilize the Douglas-Peucker algorithm with a tolerance $\epsilon = 1.2$ pixels. At our grid resolution ($r=0.044$ m/px), this corresponds to a maximum physical error bound of $\approx 5.3$ cm. This threshold was chosen to be significantly smaller than the typical robot radius, ensuring that the simplification does not affect collision checking while reducing vertex count by over 90\% (e.g., from 228k to 19k points).
    \item \textbf{Artifact Removal:} We apply an iterative spike-removal filter to eliminate rasterization artifacts. Vertices forming acute angles ($<15^{\circ}$) with short protrusions ($<0.3$m) are pruned. Crucially, vertices corresponding to passage endpoints are explicitly locked (preserved) during this process to maintain topological correctness.
\end{enumerate}

\subsubsection{OSM Serialization \& Geo-Referencing}
The final phase serializes the refined graph into the OSM XML format. Unlike generic graph formats, we ensure full geospatial compatibility by anchoring the CAD origin to a designated WGS84 coordinate (Latitude, Longitude). All metric coordinates $(x, y)$ are transformed into geodetic offsets.
In the resulting file:
\begin{itemize}
    \item Navigable spaces are encoded as closed \texttt{ways} tagged with \texttt{indoor=room}.
    \item Topological edges (passages) are encoded as \texttt{ways} connecting two room nodes, tagged with \texttt{osmAG:type=passage}.
\end{itemize}
This standardization allows the map to be immediately visualized in JOSM or parsed by osmAG path planning modules~\cite{feng2023osmag}.

\subsection{Optional Semantic Text-to-Room Association}

While geometric topology ensures navigability, semantic information enables high-level human-robot interaction (e.g., ``Go to the Kitchen''). Our pipeline treats semantic association as an optional enrichment module. In the absence of standardized annotations, the system automatically assigns unique identifiers; otherwise, we employ a heuristic score-based matching algorithm to resolve ambiguities.

Formally, let $p$ be the text annotation coordinate and $P$ be a candidate room polygon with centroid $c$ and area $A$ (in $m^2$). The assignment score is calculated using the characteristic radius $S = \sqrt{A/\pi}$, the Euclidean distance to centroid $d_c = \|p-c\|_2$, the minimum distance to boundary $d_b = \operatorname{dist}(p, \partial P)$, and the normalized centrality $\rho = d_c / S$. Matching is governed by hyperparameters $\rho_{\max}$ (default 0.7) and a physical proximity threshold $D_{\max}$ (default $\approx 2.2$m).

\textbf{Case 1: Inside Matching ($p \in P$)}
\newline
If the text annotation is located inside the polygon, the score is calculated based on its centrality. A higher score is given to text closer to the centroid.
\begin{equation}
\label{eq:score_inside}
\text{score}_{\text{inside}}(p, P) =
\begin{cases}
    100 - 50 \cdot \rho,                   & \text{if } \rho \le \rho_{\max} \\
    50 - 25 \cdot (\rho - \rho_{\max}),       & \text{if } \rho > \rho_{\max}
\end{cases}
\end{equation}

\textbf{Case 2: Nearby Matching ($p \notin P \land d_b < D_{\max}$)}
\newline
If the text is outside the polygon but within the distance threshold $D_{\max}$, the score is computed as a weighted sum of a base score, a size factor, and a distance factor. This heuristic intentionally favors associating labels with smaller rooms (e.g., offices) rather than adjacent large corridors, as validated by architectural annotation norms.
\begin{equation}
\label{eq:score_nearby}
\text{score}_{\text{nearby}}(p, P) = 40 + 30 \cdot f_{\text{size}} + 20 \cdot f_{\text{dist}}
\end{equation}
where the contributing factors are:
\begin{align}
    f_{\text{size}} &= \frac{1}{1 + \log_{10}(1 + A/10000)} \\
    f_{\text{dist}} &= 1 - \frac{d_b}{D_{\max}}
\end{align}
If neither of these conditions is met for a given text-polygon pair, the score is considered zero, and the pair is not a candidate for matching.

\textbf{Matching Strategy}: For each text entity, the algorithm computes a score against all room polygons. The room that yields the highest non-zero score is selected as the definitive match for that text label.

\begin{figure}[h!]
    \centering
    \begin{subfigure}[b]{0.48\linewidth}
        \includegraphics[width=\linewidth]{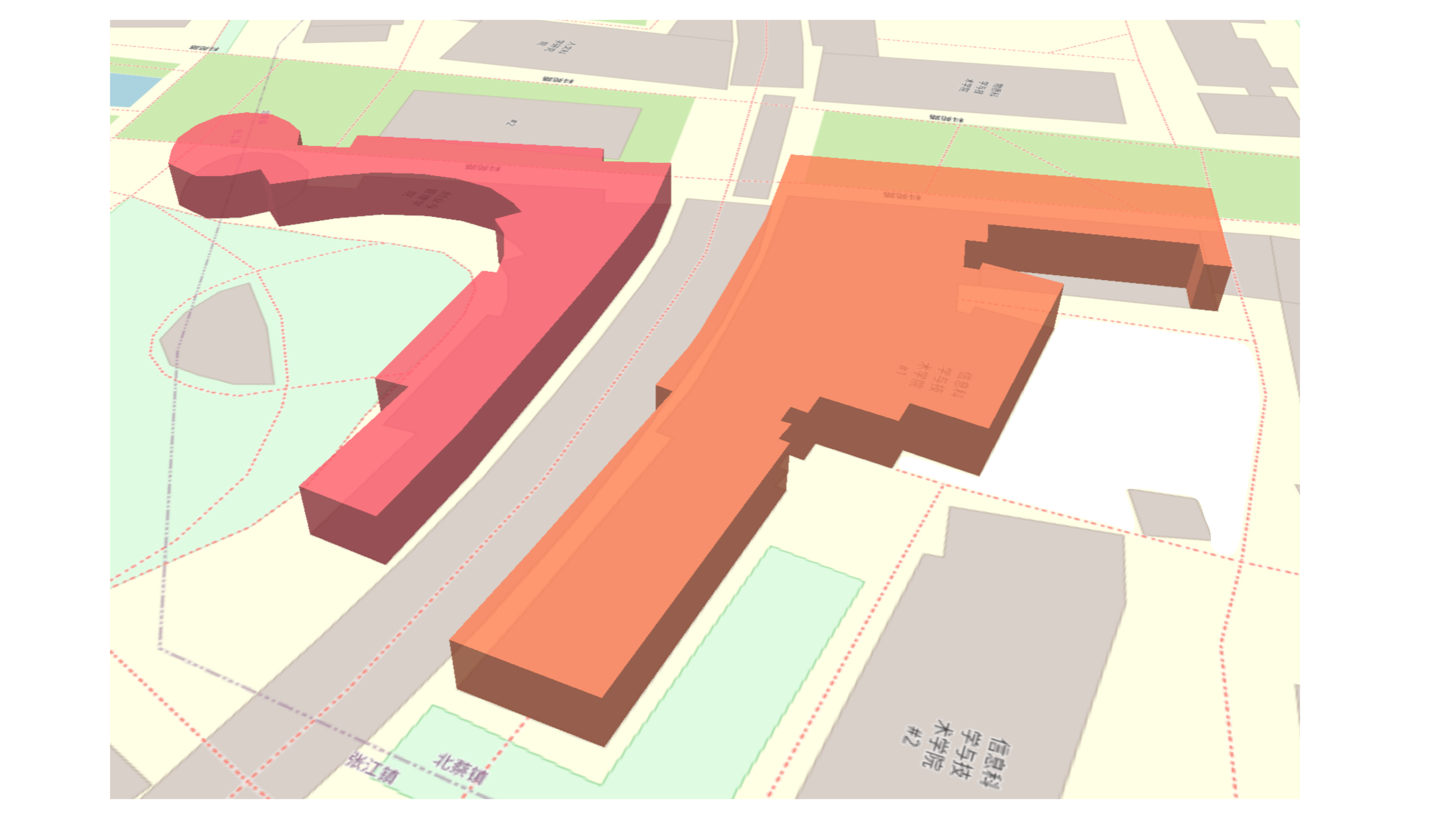}
        \caption{building level}
        \label{fig:merge1}
    \end{subfigure}
    \hfill
    \begin{subfigure}[b]{0.48\linewidth}
        \includegraphics[width=\linewidth]{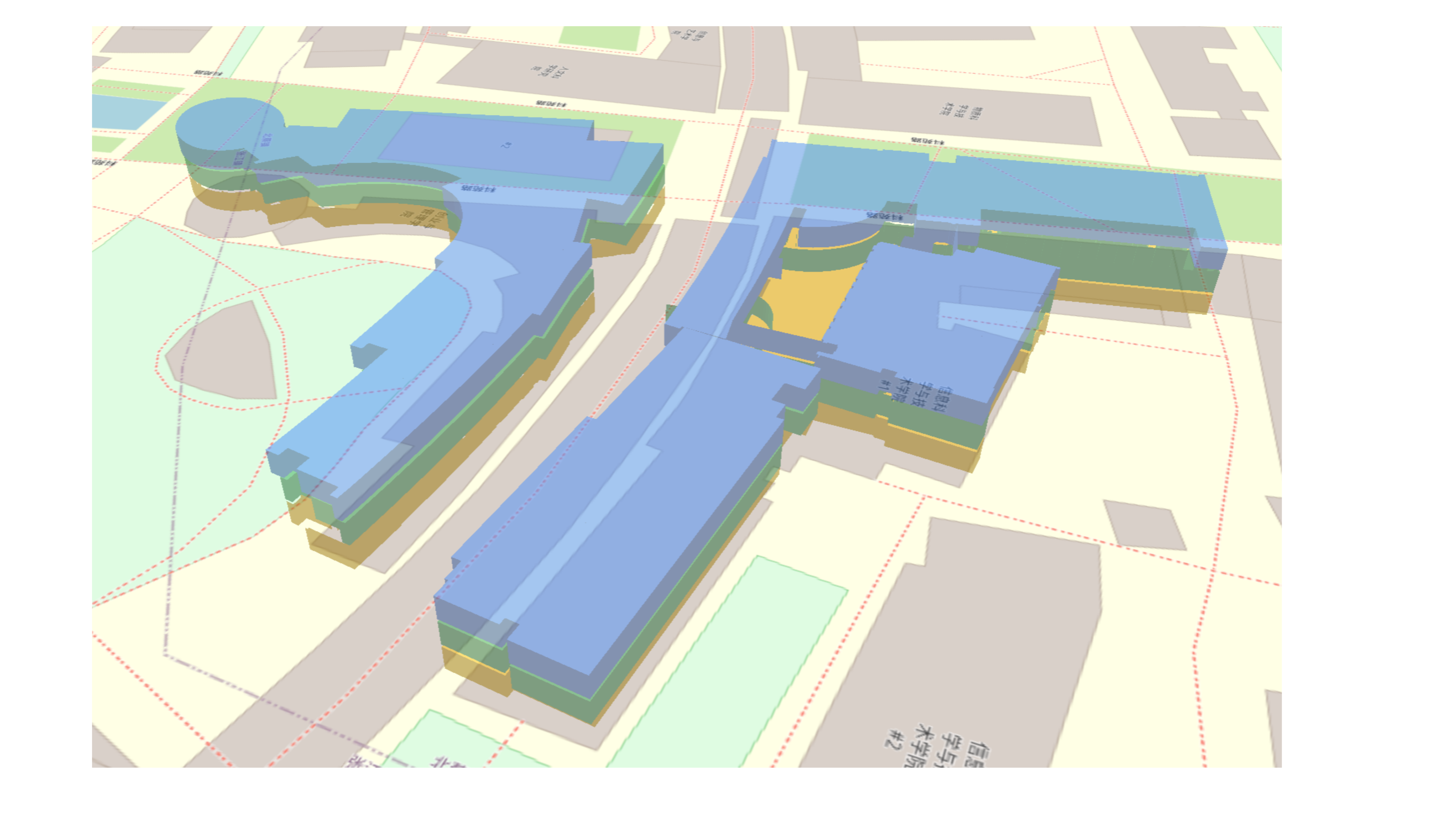}
        \caption{floor level}
        \label{fig:merge2}
    \end{subfigure}
    
    \begin{subfigure}[b]{0.48\linewidth}
        \includegraphics[width=\linewidth]{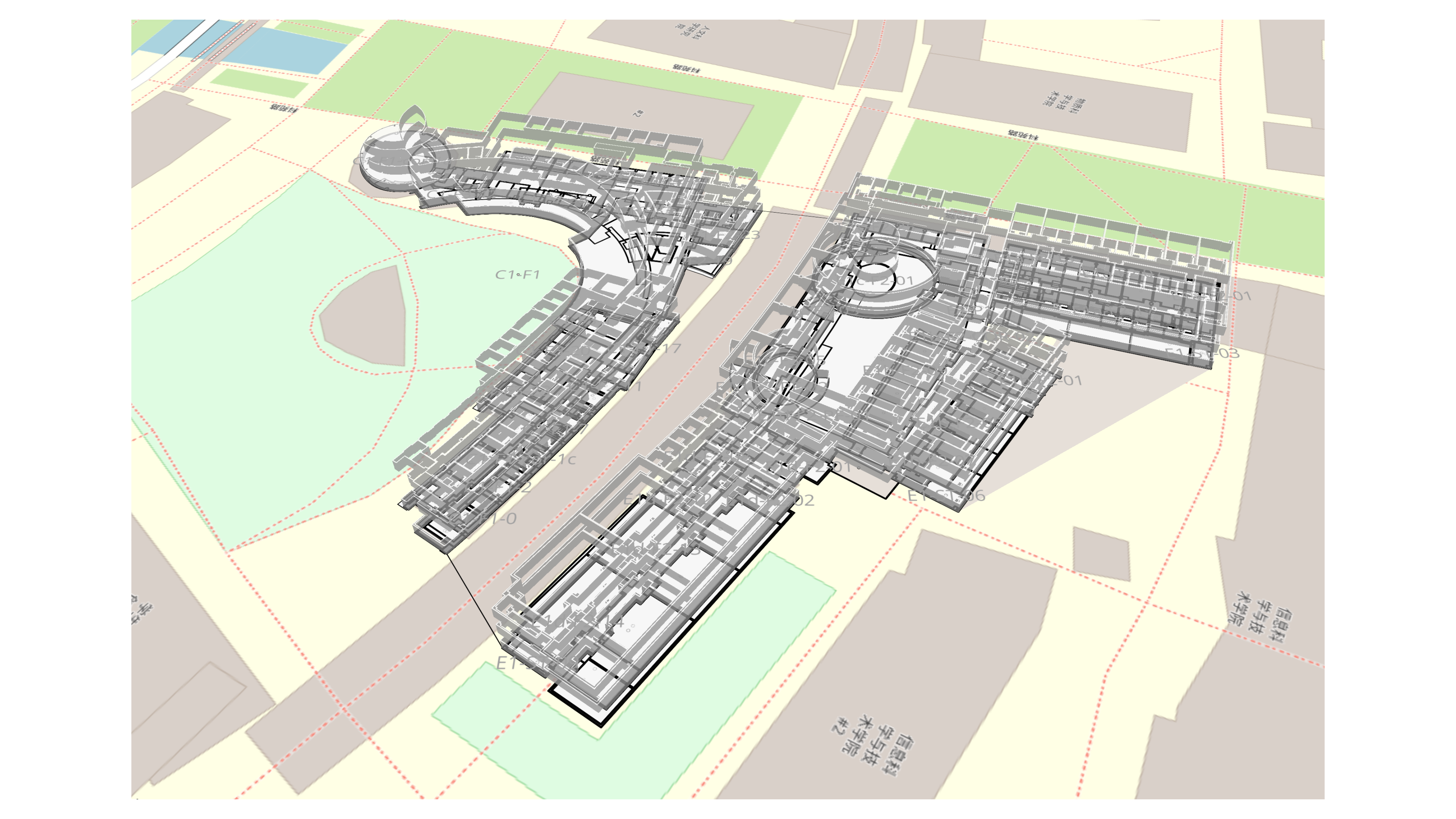}
        \caption{room level}
        \label{fig:merge3}
    \end{subfigure}
    \hfill
    \begin{subfigure}[b]{0.48\linewidth}
        \includegraphics[width=\linewidth]{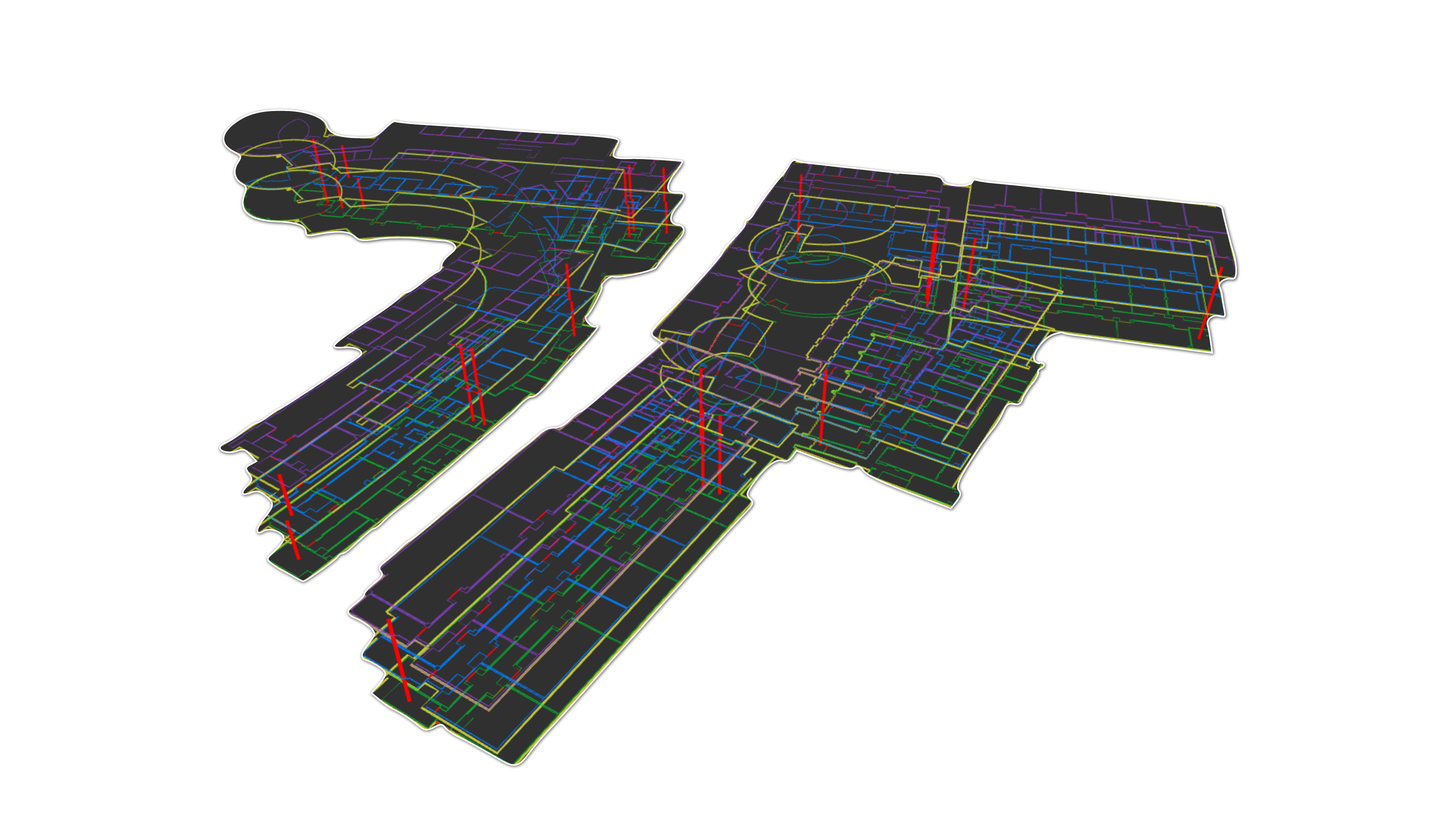}
        \caption{Rviz visualization}
        \label{fig:merge4}
    \end{subfigure}
    \caption{Results of Hierarchical Multi-Floor Map Fusion. Figures \subref{fig:merge1},\subref{fig:merge2},\subref{fig:merge3} depict three distinct hierarchical levels of a \blind{ShanghaiTech}university building as rendered in the OpenIndoor viewer\cite{osm_wiki:openindoor}. Figure \subref{fig:merge4} showcases an enhanced OSM map visualized in Rviz after being processed by osmAG parser \cite{feng2023osmag}. The red line segments represent vertical passages (e.g. elevators and stairs) that topologically connect adjacent floors.}
    \label{fig:shanghaitech-multilevel}
\end{figure}

\subsection{Hierarchical Multi-Floor Fusion}
A critical contribution of this work is the automated reconstruction of the building's 3D topology. Unlike methods that treat floors independently, we fuse them into a unified graph using a three-step algorithm:

\begin{enumerate}
    \item \textbf{Vertical Anchor Identification:} The system scans all generated maps for regions tagged with vertical transport semantics (i.e., `osmAG:areaType` is "elevator" or "stairs").
    \item \textbf{Centroid Alignment \& Registration:} We assume standard CAD files share a common coordinate origin. However, to compensate for potential layer misalignment, we implement an automatic registration step. We calculate the centroid deviation vector $\Delta \vec{v}$ between semantically identical vertical anchors (e.g., "Stair A" on Floor 1 vs. Floor 2) and apply a rigid translation to align the upper floor.
    \item \textbf{Topological Stitching:} Once aligned, if the 2D bounding boxes of two vertical transport regions on adjacent floors overlap, a new edge $E_{vertical}$ is added to the topology graph.
\end{enumerate}

This algorithm ensures topological correctness for cross-level navigation without manual edge creation.


%% file: sections/experiment.tex
\section{EXPERIMENTS}
\label{sec:experiments}

To validate the proposed pipeline, we conducted a comprehensive, bottom-up evaluation ranging from unit-level geometric fidelity to system-level robotic navigation utility. Experiments were performed on a diverse dataset of 24 architectural CAD floor plans, comprising: (1) \textbf{University Campus}: 6 multi-story plans ($\approx 11,025 m^2$) for testing large-scale fusion; and (2) \textbf{ArchWeb Public Repository}: 18 varied layouts (museums, apartments) to test generalization. Ground Truth (GT) was established via manual annotation of navigable spaces and topology. All experiments ran on a standard Intel i7 desktop, with an average processing time of \textbf{35 seconds} per floor, confirming the system's efficiency for rapid deployment.

\begin{figure}[t]
    \centering
    \includegraphics[width=0.9\columnwidth]{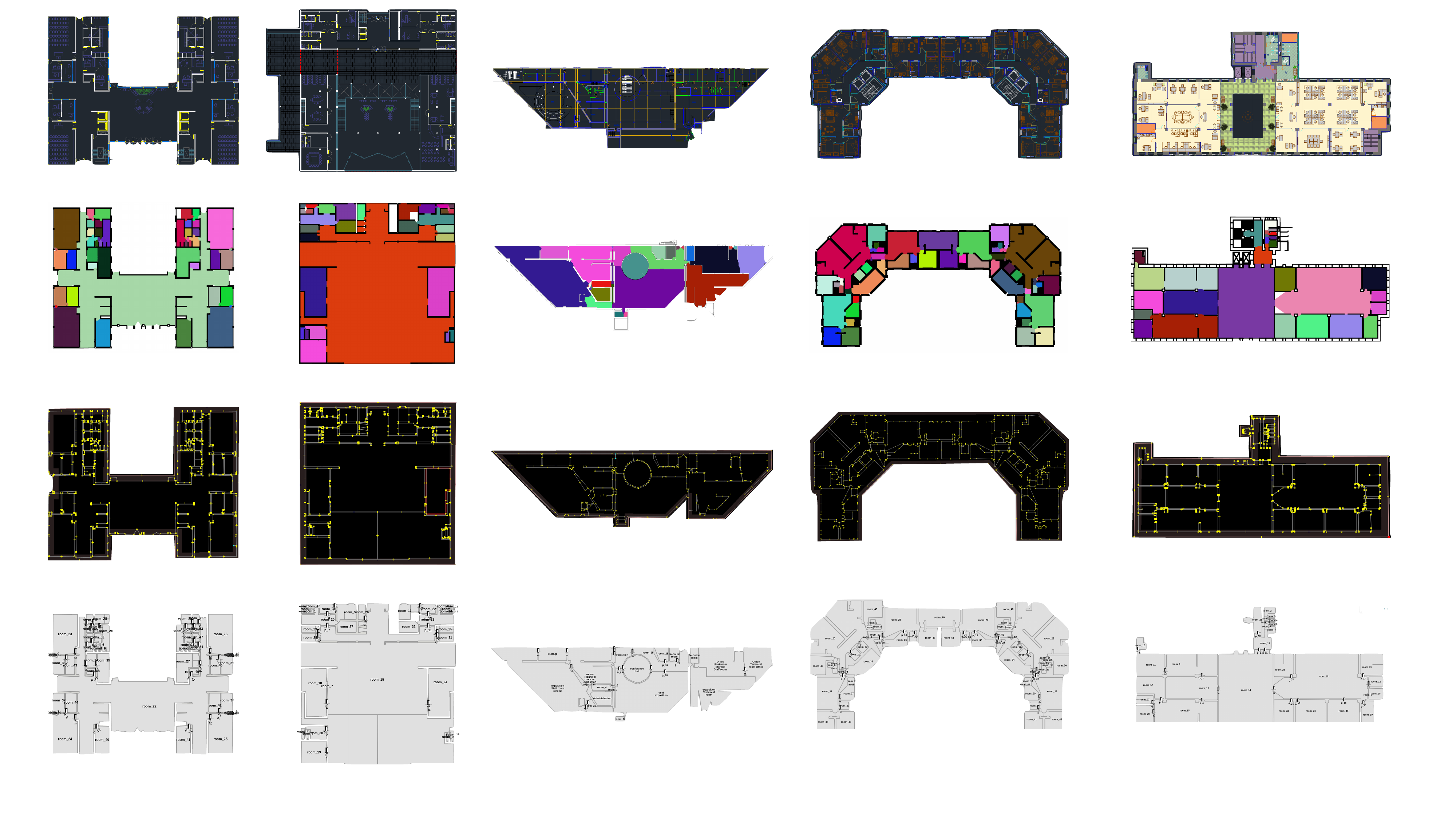} 
    \caption{Qualitative results across 5 diverse architectural styles. From top to bottom: (1) Original CAD; (2) Intermediate AreaGraph; (3) Refined OSM in JOSM; (4) Rendered 3D View. The high structural fidelity across styles confirms the method's generalization capability.}
    \label{fig:qualitative_results_gallery}
\end{figure}

\subsection{Unit-Level: Topological and Geometric Fidelity}
\label{subsec:exp_topology}

We first quantify the accuracy of the generated AreaGraph against GT. As shown in Table~\ref{tab:quantitative_results}, our pipeline achieves high Recall ($>90\%$) for both rooms and passages.
We observe a slightly lower Precision due to intentional over-segmentation (e.g., dividing large halls into sub-regions). This conservative strategy is preferred for navigation safety, as false positive boundaries are traversable, whereas false negative mergers could cause collision risks.

\begin{table}[h!]
    \centering
    \caption{Quantitative Evaluation of Segmentation (24 Maps)}
    \label{tab:quantitative_results}
    \resizebox{0.9\columnwidth}{!}{ 
        \begin{tabular}{l c c c c} 
            \toprule
            Element & GT Count & Precision & Recall & F1-Score \\
            \midrule
            Rooms      & 612 & 75.73\% & 91.71\% & 82.88\% \\
            Passages   & 647 & 77.82\% & 90.56\% & 83.70\% \\
            \bottomrule
        \end{tabular}
    }
    \vspace{-10pt} 
\end{table}

\textbf{Comparison with Deep Learning Baselines:} Although our pipeline utilizes an intermediate raster representation, direct benchmarking against Deep Learning (DL) methods~\cite{liu2017raster,zeng2019deep} is inequitable due to the supervision gap. DL models rely on domain-specific training data (e.g., R2V/R3D datasets) and fail to generalize to the clean geometry of CAD drawings without extensive retraining and manual annotation (Domain Shift). 
Quantitatively, state-of-the-art DL methods~\cite{zeng2019deep} report a \textbf{Mean IoU of $\approx$0.74} and pixel accuracy of $\approx$0.89. In contrast, our rule-based pipeline achieves superior fidelity (Recall $>\textbf{0.90}$) without requiring any training data. While a direct IoU comparison is complicated by dataset differences, our high Recall guarantees structural completeness, which is critical for collision-free planning. This highlights our key advantages: (1) Zero-Shot Generalization; (2) Determinism, guaranteeing structural validity without probabilistic hallucinations; and (3) Explainability.

\textbf{Ablation on Dynamic Parameter Adaptation:} To validate the necessity of our dynamic $\alpha$-shape formulation (Eq.~\ref{equation4alpha}), we compared it against fixed parameter baselines. We evaluated segmentation performance (F1-Score) across two distinct building types: "Narrow" (dormitories with 1.5m corridors) and "Open" (museums with $>10m$ halls).
As illustrated in Table~\ref{tab:ablation_alpha}, a fixed large $\alpha$ ($8.0m$) leads to \textbf{over-segmentation} in open spaces (erroneously splitting large halls), whereas a fixed small $\alpha$ ($2.0m$) results in \textbf{under-segmentation} in dense layouts (failing to separate narrow rooms from corridors). In contrast, our \textbf{Dynamic $\alpha$} maintains consistent high performance across both scales. This confirms the system's \textbf{scale-invariance}, effectively eliminating the need for manual parameter tuning across diverse architectural styles.

\begin{table}[h!]
    \centering
    \caption{Ablation Study: Fixed vs. Dynamic $\alpha$ (F1-Score)}
    \label{tab:ablation_alpha}
    \resizebox{0.95\columnwidth}{!}{
        \begin{tabular}{l c c c}
            \toprule
            Strategy & Narrow Layouts & Open Layouts & Avg. F1 \\
            \midrule
            Fixed $\alpha = 2.0$m & 0.32 & 0.79 & 0.55 \\
            Fixed $\alpha = 8.0$m & 0.81 & 0.45 & 0.63 \\
            \textbf{Ours (Dynamic)} & \textbf{0.86} & \textbf{0.83} & \textbf{0.85} \\
            \bottomrule
        \end{tabular}
    }
    \vspace{-10pt}
\end{table}

\subsection{Logic-Level: Semantic and Multi-Floor Integrity}
\label{subsec:exp_semantic}

\textbf{Semantic Association:} Evaluated on the 6 university campus maps containing ground truth semantic annotations, the text-to-room module achieved an average Semantic Accuracy of \textbf{91.2\%}. Failures were confined to extreme text clutter or non-standard annotations (e.g., material specs), which are handled gracefully by the system's auto-ID fallback.

\textbf{Multi-Floor Fusion and Alignment:} This algorithm operates on regions identified as vertical transport anchors (e.g., "Elevator", "Stair"). These semantic attributes are automatically derived by our text-to-room module when standard text labels are present in the CAD file; for non-compliant drawings lacking such labels, minimal manual tagging serves as a sufficient fallback to enable the automated fusion process.

Validating on a 6-story campus with 20 vertical shafts, the system established 100\% (40/40) topological connections. Our centroid-based registration yielded a \textit{Residual Alignment Error} of \textbf{0.39m}, ensuring sufficient precision for safe cross-floor navigation.

\subsection{System-Level: Navigation Verification}
\label{subsec:exp_navigation}

\begin{figure}[t]
    \centering
    \begin{subfigure}[b]{0.48\columnwidth}
        \centering
        \includegraphics[width=\linewidth]{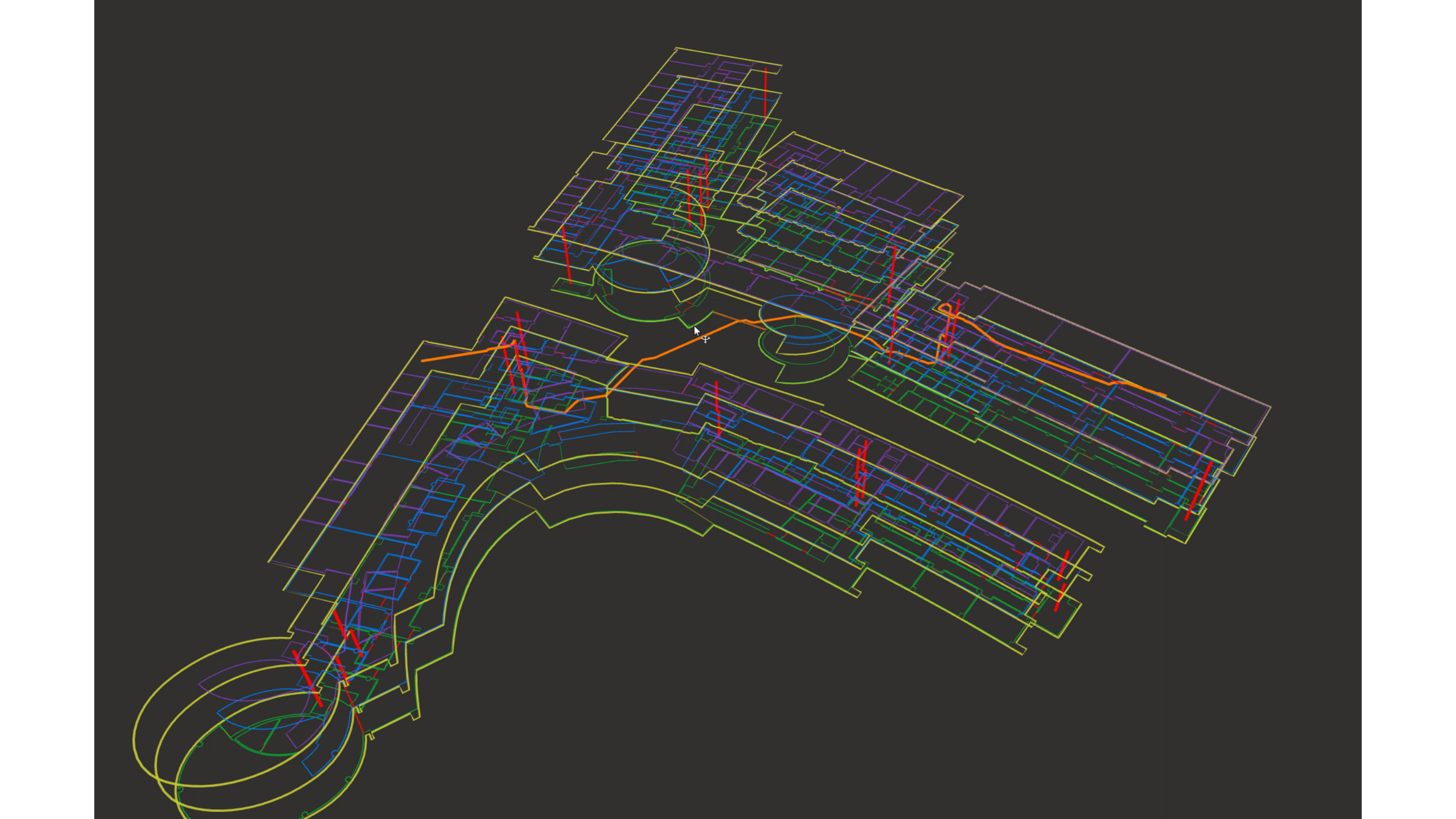} 
        \caption{Scalable Multi-floor Planning}
        \label{fig:nav_planning}
    \end{subfigure}
    \hfill 
    \begin{subfigure}[b]{0.48\columnwidth}
        \centering
        \includegraphics[width=\linewidth]{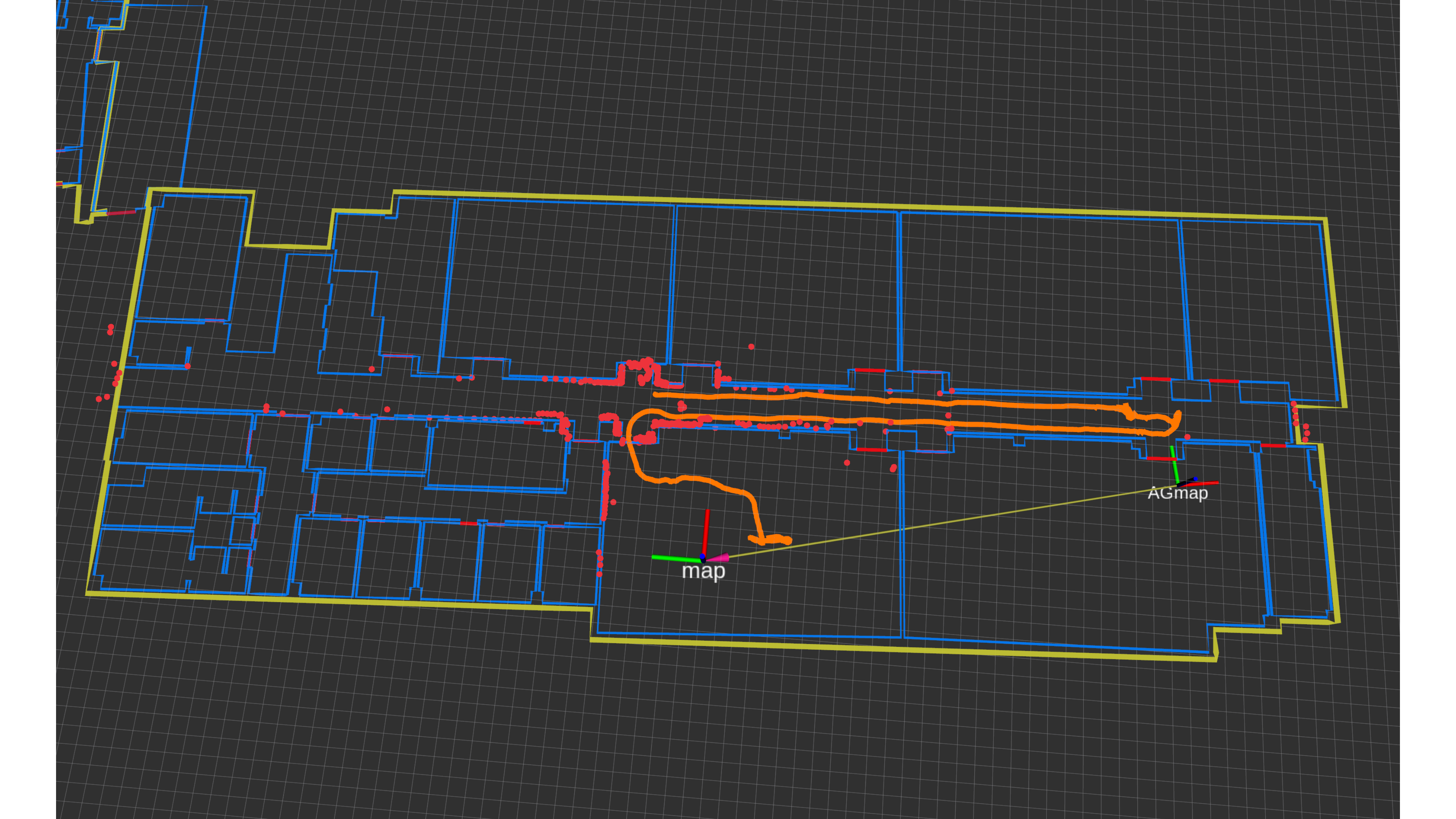}
        \caption{Robust LiDAR Localization}
        \label{fig:nav_localization}
    \end{subfigure}
    
    \caption{\textbf{System-Level Navigation Verification.} 
    (a) \textbf{Global Path Planning}: A cross-floor path computed on osmAG. The topological structure enables efficient routing ($\approx 2ms$) across levels, a capability infeasible for standard 2D grids. 
    (b) \textbf{Localization Alignment}: Real-time LiDAR scans (red) overlaid on the generated map (blue). The tight alignment validates that our geometric simplification retains sufficient metric accuracy for robust localization, demonstrating stability even within geometrically featureless long corridors.}
    \label{fig:system_verification}
\end{figure}


Crucially, we validate that the generated osmAG maps act as ready-to-use structural priors for downstream robotic tasks. We conducted independent evaluations leveraging the planning method proposed in~\cite{feng2023osmag} and the localization framework from~\cite{xie2023robust}. These experiments are designed specifically to verify the \textbf{fidelity and utility of the generated map}, utilizing these established algorithms as validation tools.

\subsubsection{Computational Tractability in Global Planning}
We benchmarked global path planning to validate the map's \textit{topological efficiency} on queries ranging from \textbf{10m to 190m}.

Results highlight two critical findings. First, the generated path lengths on our AreaGraph were nearly identical to those on the high-resolution Grid Map (e.g., for a 190m trajectory, the difference was $<0.01$m), confirming that our topological abstraction preserves geometric optimality.
Second, computational efficiency differed by orders of magnitude, especially as path length increased. While Grid A* performance degraded drastically for long-horizon queries (e.g., taking $\approx$ \textbf{19.6s} for a 190m path), our approach maintained real-time performance ($\approx$ \textbf{3.6ms}). This $\mathbf{5000\times}$ speedup confirms the system's computational scalability. Furthermore, regarding \textbf{cross-floor routing} (visualized in Fig.~\ref{fig:nav_planning}), standard 2D grid maps are inherently inapplicable due to the lack of vertical connectivity. In contrast, the osmAG naturally handles multi-story queries as a unified graph traversal, validating the system's unique capability to support navigation in complex, multi-level infrastructure.

\subsubsection{Robustness of Structural Prior Localization}
Complementing the topological verification, we assessed the \textit{geometric fidelity} of the map using the localization method from~\cite{xie2023robust}. A physical robot equipped with a 3D LiDAR executed a 200m trajectory through dynamic corridors and cluttered labs, environments significantly different from the static CAD due to pedestrians and temporary obstacles.
Despite this "Reality Gap," the generated map served as a robust structural baseline. By anchoring purely on \textit{permanent structural features} (walls, columns) extracted by our pipeline, the system achieved an \textbf{ATE RMSE of 0.18m} (Max: 0.37m) against the LIO-SAM~\cite{shan2020lio} ground truth. This confirms that our deterministic geometric simplification ($\epsilon \approx 5.3cm$) retains high-precision metric accuracy, sufficient to decouple localization stability from environmental transience.

%% file: sections/conclusion.tex
\section{Conclusion and Discussion}
\label{sec:conclusion}

In this paper, we bridge the gap between Architecture and Robotics by presenting an automated pipeline that transforms raw CAD files into Hierarchical Topometric OSM maps~\cite{feng2023osmag}. By treating architectural layouts as authoritative \textit{structural priors}, our approach circumvents the maintenance challenges of SLAM-based maps, offering a deterministic and semantically rich backbone for navigation. We validated the system's scalability and multi-floor integrity across a $11,025~m^{2}$ campus, demonstrating that our rule-based processing achieves the metric fidelity required for precise localization~\cite{xie2023robust} and global path planning~\cite{feng2023osmag}. While our dynamic parameterization ensures geometric robustness, the current reliance on standardized CAD layers remains a limitation when processing non-compliant ("spaghetti") drawings. Future work will address this by leveraging Large Language Models (LLMs) to parse non-standard layer semantics and explore a \textit{hybrid mapping paradigm}—fusing this static architectural baseline with real-time sensory updates—to enable truly persistent, life-long robot autonomy.